\documentclass{article}
\usepackage[utf8]{inputenc}
\usepackage{graphicx}
\usepackage{url}
\usepackage{float}
\usepackage{caption}
\usepackage{hyperref}
\usepackage{multirow}
\usepackage{placeins}
\usepackage{booktabs}
\usepackage{array}
\usepackage{tcolorbox}
\usepackage{verbatim}
\usepackage{amsmath}
\usepackage{amssymb}
\usepackage{natbib}
\usepackage{hyperref}
\hypersetup{
    colorlinks=true,
    linkcolor=blue,
    urlcolor=blue,
    citecolor=black,
}
\setcitestyle{authoryear,open={(},close={)}} %Citation-related commands
\usepackage[margin=1.2in]{geometry}
\title{AutoRAG: Automated Framework for optimization of Retrieval
Augmented Generation Pipeline}

\author{
    \begin{tabular}{ccc}
        Dongkyu Kim\thanks{These authors contributed equally to this work. \\Special thanks to Shivay Nagpal and Alexander Fred-Ojala for their helpful feedback and thoughtful suggestions.} & 
        Byoungwook Kim\footnotemark[1] & 
        Donggeon Han\footnotemark[1] \\
        \texttt{Markr}&\texttt{Markr}&\texttt{Markr}\\
        \texttt{\href{mailto:jeffrey@markr.ai}{jeffrey@markr.ai}} & 
        \texttt{\href{mailto:bwook@markr.ai}{bwook@markr.ai}} & \texttt{\href{mailto:eastsidegunn@markr.ai}{eastsidegunn@markr.ai}} \\
        \\
        Matouš Eibich\\
        \texttt{Predli}\\
        \texttt{\href{mailto:matous.eibich@datera.cz}{matous.eibich@datera.cz}}
    \end{tabular}
}

\begin{document}
\maketitle

\begin{abstract}
Using LLMs (Large Language Models) in conjunction with external documents has made RAG (Retrieval-Augmented Generation) an essential technology. Numerous techniques and modules for RAG are being researched, but their performance can vary across different datasets. Finding RAG modules that perform well on specific datasets is challenging.

In this paper, we propose the AutoRAG framework, which automatically identifies suitable RAG modules for a given dataset. AutoRAG explores and approximates the optimal combination of RAG modules for the dataset.

Additionally, we share the results of optimizing a dataset using AutoRAG. All experimental results and data are publicly available and can be accessed through our \href{https://github.com/Marker-Inc-Korea/AutoRAG_ARAGOG_Paper}{GitHub repository}.

\end{abstract}

\section{Introduction} % Change it to less AI-y
Large Language Models (LLMs) have significantly advanced the field of natural language processing (NLP), enabling applications from text generation to question answering. However, optimizing the integration of dynamic, external information remains challenging. Retrieval Augmented Generation (RAG) techniques address this by incorporating external knowledge sources into the generation process, enhancing the contextual relevance and accuracy of LLM outputs. 

While RAG has proven successful, the process of selecting individual RAG techniques is often not automated or optimized, limiting the potential and scalability of this technology. This lack of systematic automation leads to inefficiencies and prevents the comprehensive exploration of RAG configurations, resulting in suboptimal performance.

AutoRAG aims to bridge this gap by introducing an automated framework that systematically evaluates numerous RAG setups across different stages of the pipeline. AutoRAG optimizes the selection of RAG techniques through extensive experimentation, similar to AutoML practices in traditional machine learning. This approach streamlines the evaluation process and improves the performance and scalability of RAG systems, enabling more efficient and effective integration of external knowledge into LLM outputs.

\begin{figure}[H] \centering 
\includegraphics[width=\linewidth]{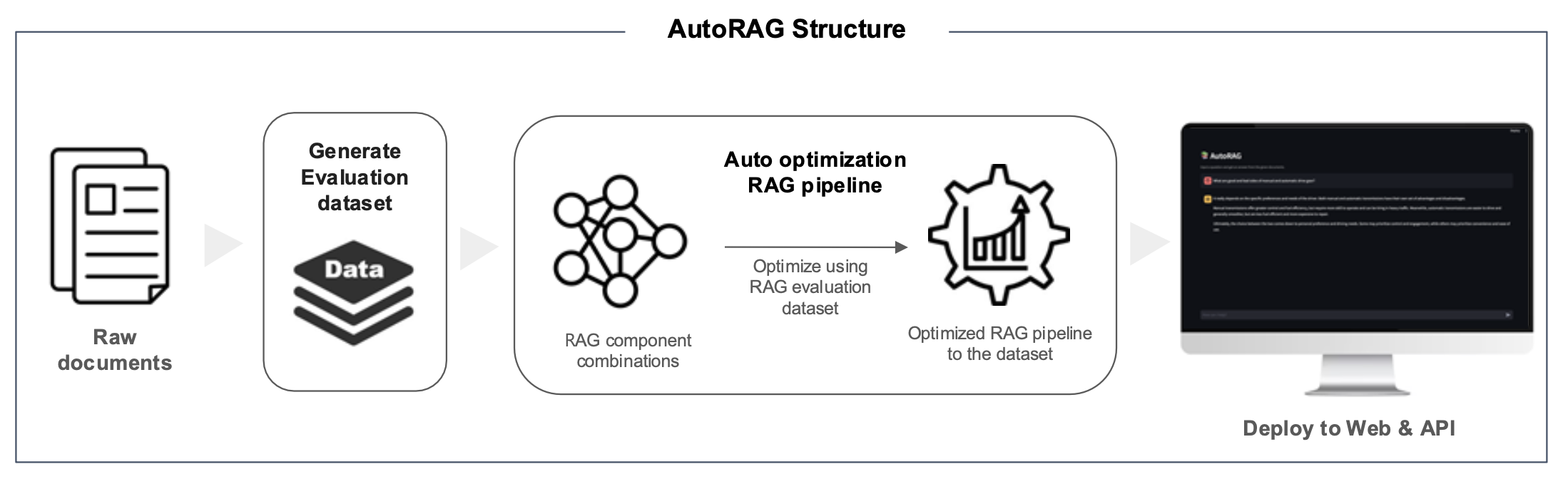} \caption{Structural diagram showing the overall structure of AutoRAG.} 
\label{fig:full-structure} 
\end{figure}

\section{RAG Techniques}
\begin{figure}[H] \centering 
\includegraphics[width=\linewidth]{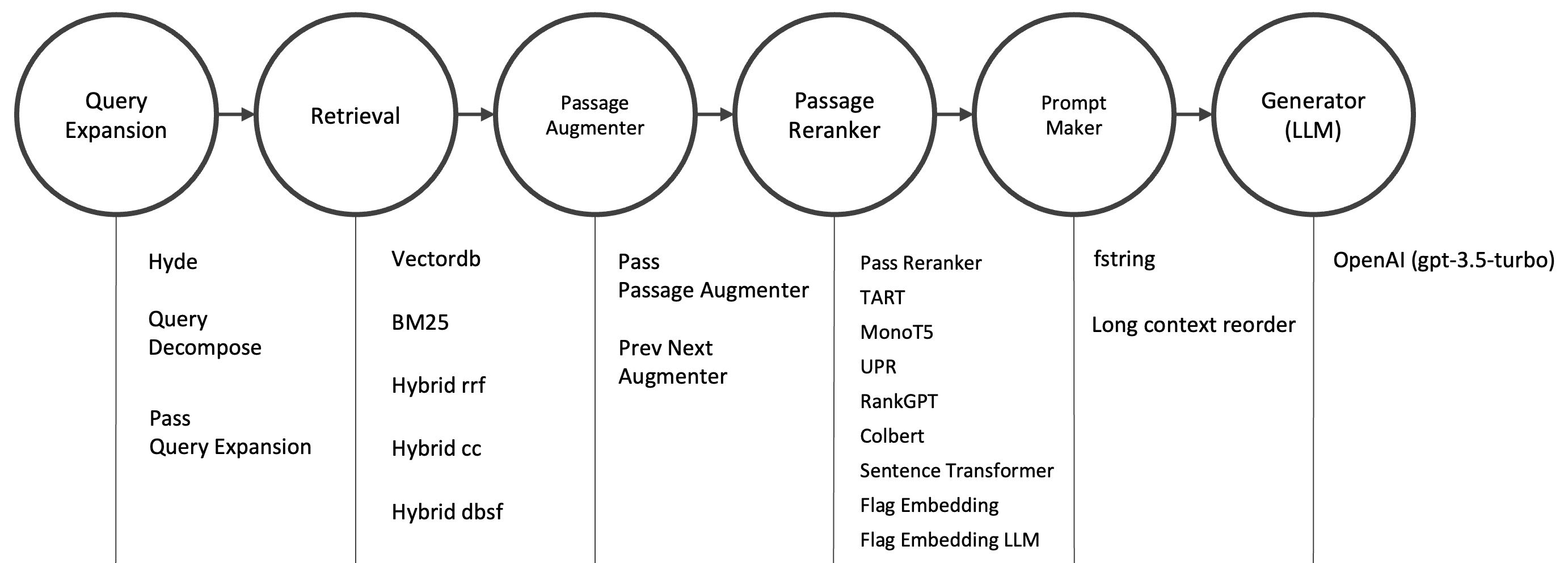} \caption{All RAG techniques used in this paper} 
\label{fig:rag-modules} 
\end{figure}

This section explores various RAG techniques evaluated in our study. We examine strategies for query expansion, retrieval, passage augmentation, passage reranking, and prompt creation. Each technique is aimed at optimizing the integration of external knowledge sources into the generation process to enhance the relevance and accuracy of LLM outputs. See figure \ref{fig:rag-modules} to check out all RAG techniques used in this paper.

\subsection{Query Expansion}
It is common to use the user's query directly as a search query in the retrieval system. However, augmenting the query can help enhance retrieval performance. A query expansion module modifies the user's query to create a better search query, making it easier to find the right passage.
\subsubsection{Query Decompose} % multi-hop question is a scientific term, so we think it is okay.
The query decomposition process is used to break down a multi-hop question into single-hop questions using a LLM. This process leverages a default decomposition prompt inspired by the StrategyQA few-shot prompt from (\cite{visconde}).

For instance, consider the multi-hop question: "What is the capital of the country where the inventor of the telephone was born?"

The query decomposition module would break this down into the following single-hop questions:

1. "Who invented the telephone?"

2. "Where was the inventor of the telephone born?"

3. "What is the capital of that country?"

By decomposing the original question into simpler queries, the retrieval system facilitates more accurate passage retrieval.
\subsubsection{HyDE}
The Hypothetical Document Embedding (HyDE) (\cite{hyde}) technique improves document retrieval by utilizing LLMs to generate a hypothetical passage to a query. This approach increases the embedding vector similarity that the hypothetical passage will be semantically similar to the actual relevant passage more than the user's query.
For instance, consider the following example:

Question: What is Ars-HDGunn structure?

HyDE expanded query: The Ars-HDGunn structure is a specialized architectural design that integrates advanced materials and innovative engineering techniques to create a highly efficient and sustainable building. This structure is characterized by its unique combination of aesthetic appeal and functional performance, often incorporating features such as green roofs, solar panels, and energy-efficient systems.

\begin{figure}[H] \centering \includegraphics[width=\linewidth]{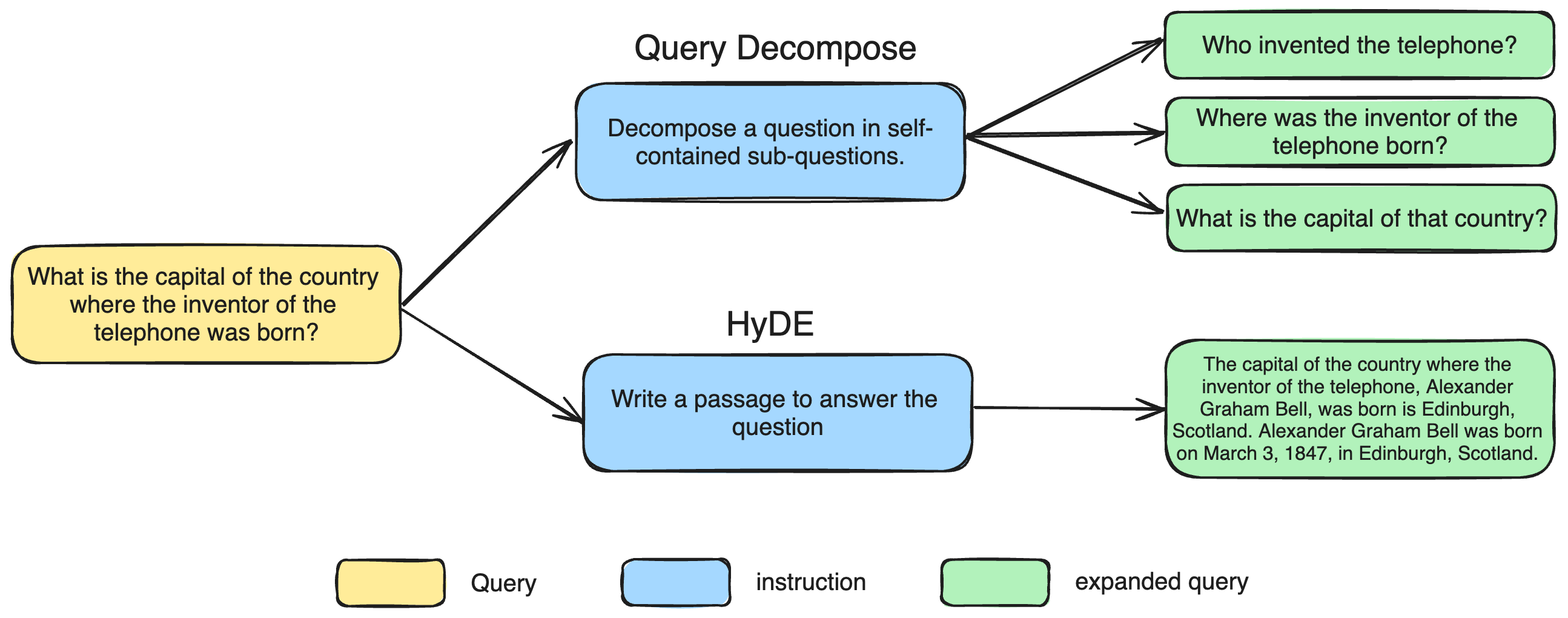} \caption{An illustration of Query Decompose and HyDE query expansion modules.} \label{fig:query-expansion} \end{figure}

% add visuals for query expansion
\subsection{Retrieval}
\subsubsection{Vectordb}
For retrieval with vector DB, passage embedding vectors are generated using a pre-trained embedding model. (\cite{dense-passage-retrieval}) These vectors represent the passages in a high-dimensional space.
Subsequently, a query embedding vector is created using the same embedding model. The semantic similarity between the query embedding and each passage embedding is then computed.

The final step involves identifying the passage embedding that has the highest similarity score with the query embedding. This approach, known as cosine similarity search, efficiently retrieves the most relevant passages by leveraging dense vector representations and similarity computations.
\subsubsection{BM25} % Might be add picture for bm25
BM25 (Best Matching 25) (\cite{bm25}) is an information retrieval module based on the probabilistic relevance framework. It extends the classic TF-IDF (Term Frequency-Inverse Document Frequency) model by introducing term frequency saturation and document length normalization. BM25 scores are calculated using a set of formulas that account for the term frequency, inverse document frequency, and the length of documents.

In a nutshell, it uses words in queries for searching documents. The more words in a query appear in the document, the more relevance scores it gets. While searching, it uses only infrequent words in the document, ignoring frequent words. Because frequent terms like 'you', 'and', or 'like' can be irrelevant to the query and document meaning.

In this paper, we used 'rank-bm25' library(\cite{rankbm25}) for BM25 calculation.

\subsubsection{Hybrid Retrievals}

Hybrid retrieval is the fusion of sparse retrieval methods like BM25 and dense retrieval methods like vector databases that use embedding models. In this paper, we used four hybrid retrieval variants. Using hybrid retrieval can enhance performance because it leverages both the lexical similarity of sparse retrieval and the semantic similarity of dense retrieval.

\textbf{Hybrid RRF} (\cite{rrf}) uses the reciprocal rank fusion algorithm to fuse results.

RRF can be represented by the following formula:

$$
f_{RRF}(q, d) = \frac{1}{\eta + \pi_{LEX}(q,d)} + \frac{1}{\eta + \pi_{SEM}(q,d)}
$$

The $\pi(q,d)$ is the rank function, which gets the rank of the document $d$ to the given query $q$ in the set of documents. In other words, the document $d$ is a subset of the retrieved document set to the query $q$.
The $\pi_{LEX}(q,d)$ is the rank of the passage in the lexical retrieval, which is BM25 in this paper.
The $\pi_{SEM}(q, d)$ is the rank of the passage in the semantic retrieval, which is 'vectordb' in this paper.
The $\eta$ is a free parameter. 

\textbf{Hybrid CC} (\cite{hybrid-cc}) and \textbf{Hybrid DBSF} use convex combination to fuse sparse and dense retrieval. Both methods use a convex combination for fusing, but the difference lies in the normalization method. The Hybrid CC retrieval uses min-max normalization, whereas the Hybrid DBSF uses the 3-sigma value as the min and max values in normalization. Both retrieval methods can be represented by the following formula:

$$
f_{Convex}(q, d) = \alpha\phi_{LEX}(f_{LEX}(q, d)) + (1 - \alpha)\phi_{SEM}(f_{SEM}(q, d))
$$

Here, $f_{LEX}(d)$ is the lexical retrieval relevance score of the document, which is the BM25 relevance score in this paper. $f_{SEM}(d)$ is the semantic retrieval relevance score of the document, which is the vectordb relevance score in this paper. The $\phi_{LEX}$ and the $\phi_{SEM}$ are the normalized functions of each retrieval method. The range of the lexical retrieval score and the semantic retrieval score is different, so the normalization process is crucial.
The $alpha$ is the weight parameter and must be in the range of 0 to 1.

The min-max normalization process for hybrid cc can be represented by the following formula:

$$
\phi_{MM}(f_o(q,d)) = \frac{f_o(q,d) - m_o(q)}{M_o(q) - m_o(q)}
$$

Here, $m_o(q)$ is the minimum relevance score value of the given retrieval method, and $M_o(q)$ is the maximum relevance score value of the given retrieval method.
In other words, the semantic and lexical normalization uses different min and max values in this paper.

The 3-sigma normalization used in hybrid DBSF retrieval can be represented by the following formula:

$$
\phi_{ts}(f_o(q,d)) = \frac{f_o(q, d) - (\mu_o - 3\sigma_o)}{(\mu_o + 3\sigma_o) - (\mu_o - 3\sigma_o)}
$$

In this formula, the $\mu_o$ is the mean value of the given retrieval method. And the $\sigma_o$ is the standard deviation value of the given retrieval method.

\subsection{Passage Augmenter}
The passage augmenter is a technique designed to enhance the performance of retrieval by obtaining additional relevant passages. This method expands the initial set of retrieved passages, thereby increasing the comprehensiveness and relevance of the information retrieved.

The process begins with an initial retrieval phase where a set of passages is obtained based on a query. Following this, the passage augmenter utilizes the metadata associated with these passages—such as author information, publication date, and keywords to perform a secondary search. In this paper, we only use metadata about neighboring passages. The secondary search aims to find more passages that are contextually related to the initially retrieved set. 

\subsubsection{prev next augmenter}
The Prev-Next Passage Augmenter designed to enhance the retrieval performance by incorporating neighboring passages.

During the chunking process, each passage can be annotated with its preceding and succeeding passages. The Prev-Next Passage Augmenter leverages this structural information to retrieve not only the initially relevant passages but also their neighbors. This approach is based on the hypothesis that adjacent passages may contain additional context or relevant information that can improve the overall retrieval performance.

\subsection{Passage Reranker} % will make a brief image that explains each type of passage rerankers
The passage reranker is a component in information retrieval systems, tasked with re-ranking passages after the initial retrieval phase. this method, while computationally expensive, has been suggested to enhance accuracy of retrieval modules(\cite{lin2019theneuralhypeandcomparisonsagainstweakbaselines}).

In the context of RAG, the passage reranker plays a vital role. RAG are often constrained by the token limit and the high computational cost of LLM. By employing a passage reranker, the system can achieve higher accuracy in identifying the most relevant passages to the query, ensuring efficient use of prompt tokens.

\begin{table}[H]
\centering
\begin{tabular}{|c|c|}
\hline
\textbf{Passage Reranker}     & \textbf{Type}             \\ \hline
MonoT5                        & \multirow{5}{*}{LM-based} \\ \cline{1-1}
Sentence Transformer Reranker &                           \\ \cline{1-1}
Flag Reranker                 &                           \\ \cline{1-1}
Flag LLM Reranker             &                           \\ \cline{1-1}
TART                          &                           \\ \hline
RankGPT                       & LLM-based                 \\ \hline
Colbert                       & Embedding-based           \\ \hline
UPR                           & Log prob-based            \\ \hline
\end{tabular}
\caption{The passage rerankers and its type we used in this paper}
\label{Table 2}
\end{table}

\begin{figure}[H] \centering \includegraphics[width=\linewidth]{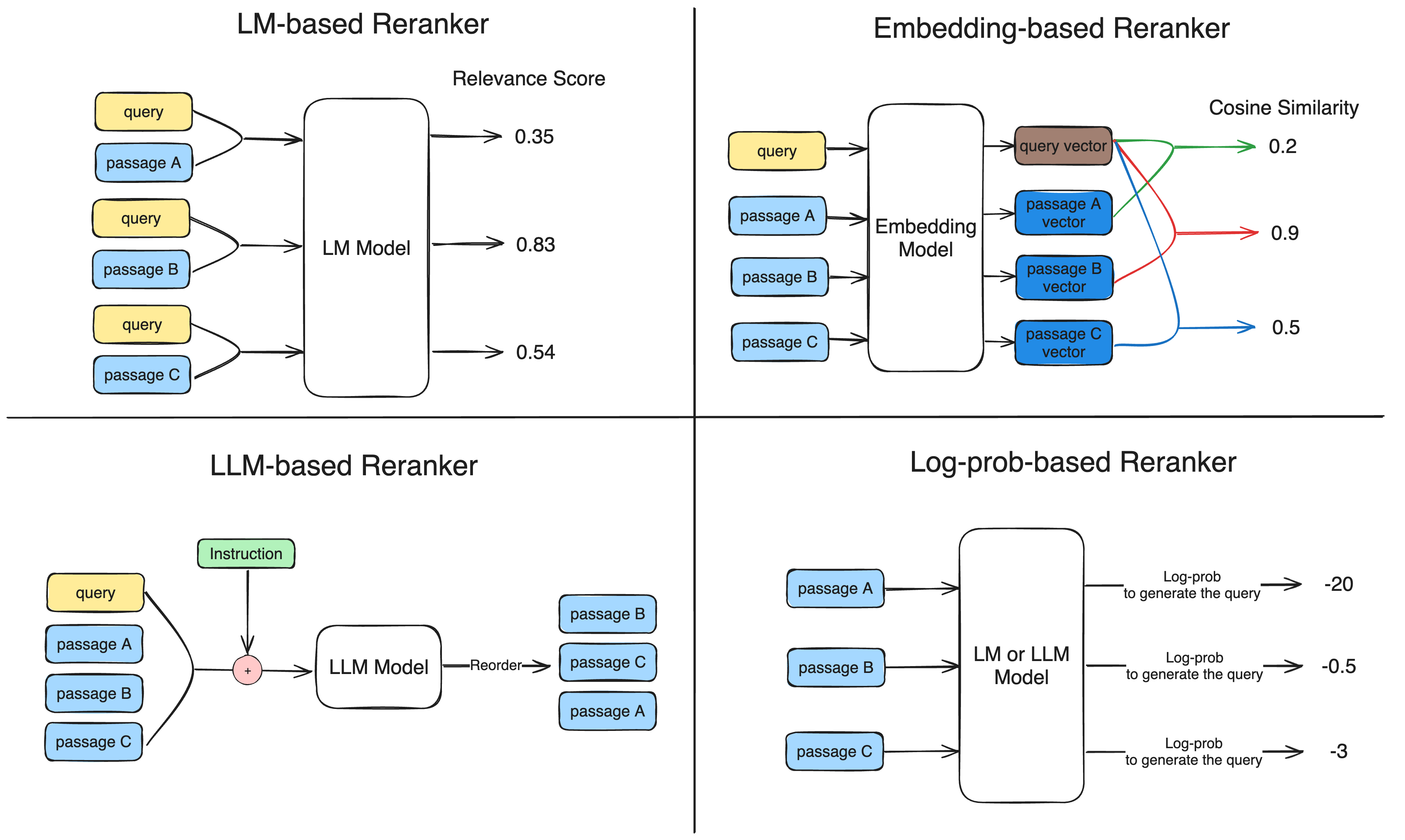} \caption{An illustration of each passage reranker module.} \label{fig:reranker} \end{figure}

\subsubsection{LM-based Reranker}
LM-based rerankers utilize fine-tuned language models to score the relevance of query-passage pairs. The training dataset comprises query-passage pairs annotated with relevance labels. The training process involves:

For relevant query-passage pairs, the model is trained to output the token 'True'.
For non-relevant pairs, the model is trained to output the token 'False'.
During inference, the model calculates the probability of outputting the 'True' token. This probability is used as the relevance score for the passage.

\textbf{MonoT5 Reranker} (\cite{monot5}) uses the T5 model (\cite{t5}) and is trained with query-passage pairs labeled for relevance. In this paper, the model is based on the T5-3B variant and is fine-tuned using the MS MARCO dataset (\cite{msmarco}) over 10,000 steps (equivalent to 1 epoch).

The \textbf{Sentence Transformer Reranker} employs a cross-encoder model to rerank retrieved passages. In this paper, the ms-marco-MiniLM-L-2-v2 model is utilized for this purpose. This model is fine-tuned to classify query-passage relevance using the MSMARCO dataset (\cite{msmarco}), which is based on the BERT model.

\textbf{Flag reranker} uses a BGE model to rerank passages. This model is based on xlm-roberta-base model(\cite{DBLP:journals/corr/abs-1911-02116}) and fine-tuned using multilingual datasets like Mr.Tydi dataset (\cite{mrtydi}).

\textbf{Flag llm reranker} is based on a LLM that has many weights compared to other rerankers. In this paper, it uses gemma-2b model (\cite{gemma}), fine-tuned for reranker usage. 

The \textbf{TART reranker} (\cite{tart}) is an LM-based reranker but uses task-specific instructions in the reranking process. 
TART lies in its ability to include instructions during the reranking phase, thereby capturing the user's intent beyond the explicit query. Other rerankers rely solely on the user's query, which may not fully encapsulate the user's underlying intent. TART addresses this limitation by allowing the inclusion of additional instructions that specify the user's intent.

The major distinction between TART and other rerankers lies in the training methodology. Other LM-based rerankers are trained on query-passage pairs with relevance labels. In contrast, TART trains the reranker model using an instruction-query-passage set. This approach allows TART to understand and incorporate the user's intent, as specified by the instructions, into the reranking process.

\subsubsection{LLM-based Reranker} % add image for LLM-base Reranker
LLM-based rerankers leverage LLMs to reorder passages based on their relevance to a given query. Unlike other rerankers that require fine-tuning the language model, LLM-based rerankers utilize prompt engineering to achieve passage reranking.

\textbf{RankGPT} (\cite{rankgpt}) inputs the query and the passages to be reranked into the LLM, which then generates permutations of the passages. These generated permutations are the reranked order of the passages.
\subsubsection{Embedding-based Reranker}
Embedding-based rerankers leverage dense vector representations to capture semantic similarities between queries and documents. For reranking passages, they generate embedding vectors for both queries and passages and then calculate the similarity of each vector. Embedding-based rerankers can be an ensemble of different embedding models. For better results, it is common to choose an embedding model different from the one used in the retrieval phase.

\textbf{ColBERT reranker} (\cite{khattab2020colbert}) independently encoding queries and passages with BERT. Then it calculates senmantic simliarity using encoded vectors.
ColBERTv2 (\cite{santhanam2022colbertv2}) uses lightweight token representations, optimizing computation cost and maintaining rerank effectiveness. We employ ColBERTv2 in this paper.
\subsubsection{Log prob-based Reranker}
Log-probability based rerankers leverage the log probability of generating a query from a given passage to assess relevance. If the log probability of generating a query is higher, the passage is more relevant.

The \textbf{UPR} (Unsupervised Passage Reranker) (\cite{sachan2023improving}) exemplifies this approach by utilizing a pre-trained language model to compute the log probability of generating a query at the given passage. In this paper, we use T5-large model(\cite{2020t5}) for UPR reranker.

\subsection{Prompt Maker}
For in-context learning, the retrieved passages must be included in the prompt to the LLM. Prompt maker is the module that concatenates retrieved passage contents, user queries, and instructions. 
\subsubsection{f-string}
This module concatenates the user's query, retrieved passages, and instructions. The higher relevance passage will be the first, and the lowest will be the last. In this paper, the top-k was set as five, so it uses the top five relevant passages. 

\subsubsection{long context reorder}
Long context reorder addresses the 'Lost in the Middle' phenomenon (\cite{lost-in-the-middle}), where large language models (LLMs) tend to prioritize the beginning and end of input prompts, often neglecting the middle content.

To mitigate this, the long context reorder module appends the most relevant passage at the end of the input prompt, ensuring it appears both at the beginning and the end. This redundancy helps LLMs maintain focus on critical information, thereby enhancing the performance of generated responses.

\section{Experiment} \label{experiment-section}
\subsection{AutoRAG}
Despite the development of numerous RAG techniques and metrics through research, these advancements have been scattered, making it challenging to identify the appropriate RAG pipeline for real-world applications. To address this issue, we propose AutoRAG, an open-source framework designed for RAG experimentation and optimization. 
AutoRAG leverages a greedy algorithm to efficiently search for the optimal initial pipeline. By organizing the model into modular nodes, each performing distinct tasks, AutoRAG dynamically selects the most promising node at each step using a strategy defined by metrics available for each node. This approach enables AutoRAG to construct near-optimal pipelines without requiring exhaustive search methods, offering both scalability and computational efficiency across diverse machine learning tasks.

\subsubsection{Node}
A 'node' corresponds to a specific step within RAG and operates as a high-level concept that acts as a container for modules. The modules that can be placed within the same node must have the same input and output formats as the node. Except for the initial node input (User query) and the final node output (Answer), the output of one node is used as the input for the subsequent node. The concepts of nodes and modules used in this experiment are illustrated in Figure \ref{fig:rag-modules}.

\subsubsection{Strategy}
In AutoRAG, the term "strategy" refers to how we choose which module to use within a node. This involves selecting and arranging different RAG techniques(modules). Each node can use performance metrics and the time a module takes to complete its task. We can also use statistical measures like the average of all these metrics to help make decisions. By defining what makes a RAG module "good" using these performance metrics, we can compare different modules. In AutoRAG, the function that serves as the criteria for selecting and optimizing modules is called the "strategy."

\subsubsection{Optimization}
Evaluating nodes like `query\_expansion` or `prompt\_maker` based solely on their outputs is challenging. In such cases, we utilize the evaluation of the subsequent node. For instance, the output of the `query\_expansion` node is one or more queries for retrieval, which makes it difficult to evaluate the modules. The `query\_expansion` node is positioned before the `retrieval` node which is relatively easier to evaluate. Therefore, we fix the modules of the `retrieval` node and only change the modules of the `query\_expansion` node for experiments. Similarly, the `prompt\_maker` node is evaluated by fixing the modules in the `generation` node.

When the preceding node (A) is difficult to evaluate and the subsequent node (B) that is easier to evaluate, with the number of modules to evaluate being \(m\) and \(n\) respectively, a comprehensive experiment would require \(m \times n\) combinations. However, using the above method, we perform \(m\) trials at stage A and \(n\) trials at stage B, thereby reducing the number of required experiments to only $m+n$ combinations.

\subsection{Data} 
This study utilizes the ARAGOG dataset(\cite{eibich2024aragogadvancedragoutput}), a tailored dataset derived from the AI ArXiv collection, accessible via Hugging
Face (\cite{AIArXivcollection}). The dataset consists of 423 selected research papers centered around the
themes of AI and LLMs, sourced from arXiv. This selection offers a comprehensive foundation for
constructing a database to test the RAG techniques and creating a set of evaluation data to assess
their effectiveness.

\subsubsection{RAG Database Construction}
For the study, a subset of 13 research papers were selected for their potential to generate specific, technical questions suitable for evaluating Retrieval-Augmented Generation (RAG) systems. To better simulate a real-world vector database
environment, where noise and irrelevant documents are present, the database was expanded to include the full dataset of 423 papers available. The papers were chunked using a chunk size of 512 tokens and an overlap of 50 tokens.

\subsubsection{Evaluation Data Preparation}
The evaluation dataset comprises 107 question-answer (QA) pairs generated with the assistance of
GPT-4. The generation process was guided by specific criteria to ensure that the questions were
challenging, technically precise, and reflective of potential user inquiries sent to an RAG system. Each
QA pair was then reviewed by humans to validate its relevance and accuracy, ensuring that the
evaluation data accurately measures the RAG techniques’ performance in real-world applications.
The QA dataset is available in this paper’s associated \href{https://github.com/Marker-Inc-Korea/AutoRAG_ARAGOG_Paper}{Github repository}.

\subsection{Metrics}

\begin{figure}[H] \centering 
\includegraphics[width=\linewidth]{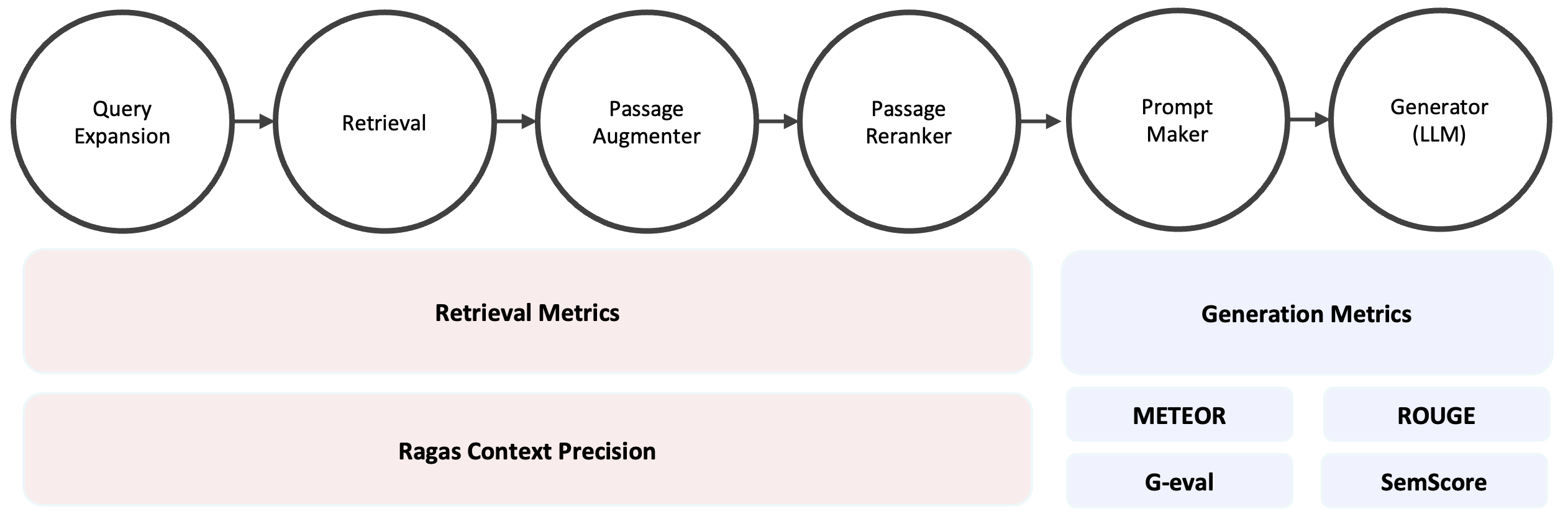} \caption{Metrics used at each stages} 
\label{fig:query-expansion} 
\end{figure}

Our experiment employs LLM-based retrieval metrics to find an appropriate retrieval structure at the Retrieval stage for QA problems that are not mapped to readily available knowledge snippets.

\subsubsection{Retrieval Metric}

Ragas (Retrieval Augmented Generation Assessment)(\cite{es2023ragasautomatedevaluationretrieval}) is a framework designed for reference-free evaluation tools. In our work, we employ the Ragas Context Precision metric.
\[Context Precision@K=\frac{\sum_{k=1}^{K}(Precision@k\times v_{k})}{true positives@K}\]
\[Precision@k=\frac{true positives@k}{(true positives@k + false positives@k)}\]
\(K\) is the total number of retrieved passages(prediction class: \(positive\)). And, \(v_{k} \in \left\{0, 1\right\}\) is the relevance indicator at rank \(k\). K is a parameter that can be set at each retrieval stage. In \ref{experiment-section}, the parameter 'top\_k' related to ragas\_context\_precision refers to this K value. We utilize GPT-4 turbo to evaluate each retrieved passage and determine its actual class—whether the passage is relevant (\(true\)) or irrelevant (\(false\)) to the given query and the target generation output.
\subsubsection{Generation Metric}

One of the challenges in evaluating the generation of outputs by large language models (LLMs) is the lack of a single metric that encompasses all perspectives. Therefore, we use a normalized mean of four metrics to determine the best generation model at each stage.

To intuitively consider the use of strong references and specialized terminology, we employ n-gram based metrics such as ROUGE and METEOR.

To account for semantic similarity with the answer, we adopt the SemScore(\cite{aynetdinov2024semscoreautomatedevaluationinstructiontuned}). Using a well-trained embedding model, we compute the cosine similarity in the embedding space between a target text and a generated text. In the experiment, OpenAI's $text-embedding-ada-002$ was employed as the embedding model for SemScore.

Additionally, we choose G-Eval(\cite{liu2023gevalnlgevaluationusing})(GPT4 turbo) to evaluate generation quality comprehensively. G-Eval is an evaluation framework utilizing a Chain-of-Thought(CoT) technique based large language model. This framework employs LLMs to generate scores ranging from 1 to 5. Through prompt engineering, it allows for evaluations from various perspectives. In this experiment, we adopted the perspectives of coherence, consistency, fluency, and relevance, using OpenAI's GPT-4 turbo (gpt-4-0125-preview) model. In this paper, we utilize the average scores for these four aspects as the g\_eval score.

\subsection{Candidate modules}

In our study, we focused on implementing and evaluating advanced RAG methods (\cite{gao2024retrievalaugmented}). The experimental setup included several distinct stages. The stages comprised Query Expansion, Retrieval, Passage Augmentation, Passage Reranking, Prompt Creation, and Text Generation. For each stage, we conducted evaluations to select the best-performing module. The output from the best-performing module would be used as the input for the subsequent stage. Below, we detail the modules tested, the metrics used for evaluation, and additional pertinent information for each stage. Note that modules named with the prefix 'pass' indicate that the module produces the same output as its input.

\subsubsection{Query Expansion}

The `top k` of ragas context precision set to 10. And the modules tested included in the Query Expansion stage:

\begin{itemize}
\item \texttt{pass\_query\_expansion} : A module that outputs the same input.
\item \texttt{query\_decompose} (LLM: OpenAI(gpt-3.5-turbo), temperature : [0.2, 1.0])
\item \texttt{hyde} (LLM: OpenAI(gpt-3.5-turbo), max\_token : 64)
\end{itemize}

\subsubsection{Retrieval}

The `top k` of ragas context precision set to 10. And the modules tested included in the Retrieval stage:

\begin{itemize}
\item \texttt{bm25} (tokenizer: GPT-2)
\item \texttt{vectordb} (OpenAI\_embed\_3\_large)
\item \texttt{hybrid\_rrf} (rrf\_k : [3, 5, 10])
\item \texttt{hybrid\_cc} (cc : [0.3, 0.7])
\item \texttt{hybrid\_dbsf} (dbsf : [0.7, 0.3]) % different weight parameter?
\end{itemize}

\subsubsection{Passage Augmenter}

The `top k` of ragas context precision set to 15. And the modules tested included in the Retrieval stage:

\begin{itemize}
\item \texttt{pass\_passage\_augmenter}
\item \texttt{prev\_next\_augmenter} (mode: both)
\end{itemize}

\subsubsection{Passage Reranker}

The `top k` of ragas context precision set to 5. And the modules tested included in the Passage Reranking stage:

\begin{itemize}
\item \texttt{pass\_reranker}
\item \texttt{tart}
\item \texttt{monot5}
\item \texttt{upr}
\item \texttt{rankgpt}
\item \texttt{colbert\_reranker}
\item \texttt{sentence\_transformer\_reranker}
\item \texttt{flag\_embedding\_reranker}
\item \texttt{flag\_embedding\_llm\_reranker}
\end{itemize}

\subsubsection{Prompt Maker}

For the Prompt Maker stage, we employed multiple metrics, including METEOR, ROUGE, and sem\_score (OpenAI), alongside `g\_eval` (GPT-4-0125-preview), averaged for comprehensive evaluation. Since a fixed module in the generator node is required to evaluate the `prompt\_maker`, we used `llama\_index` implemented with OpenAI's GPT-3.5 Turbo (temperature 0.0) as a fixed generator module. The modules tested were:
\begin{itemize}
\item \texttt{f\_string}
\item \texttt{long\_context\_reorder}
\end{itemize}

\subsubsection{Generator}

The modules tested included in the Generator stage:
\begin{itemize}
\item \texttt{llama\_index\_llm} (OpenAI GPT-3.5 Turbo, temperature 0.0)
\end{itemize}

Each of these stages were thoroughly evaluated to ensure that the best-performing modules were selected based on their respective metrics, ensuring that the subsequent stages received the most effective input possible. The following sections detail the results and analysis of each stage's experiments. Additionally, both METEOR and execution time are provided to offer more detailed evaluation results but are not used for module selection.

\section{Results}
This study systematically evaluates different advanced RAG techniques using Retrieval Metrics and Generation Metrics. A comparative analysis is presented using bar plots to visualize the distribution of these metrics, and the results are interpreted with tables.

\subsection{Retrieval Metric}

To evaluate the performance of our retrieval system, we utilized the Ragas Context Precision as the primary retrieval metric. This metric was applied across a set of 107 queries, yielding 107 individual precision scores for each experimental module. For each module, we computed the mean Ragas Context Precision score from the 107 individual scores. These mean values provide a summary measure of each module's retrieval effectiveness. 

To facilitate a comparative analysis of the modules, we visualized the mean Ragas Context Precision scores using bar plots. This visualization allows for a clear comparison of the retrieval performance across the different modules.

\subsubsection{Query Expansion}

The bar plots for Context Precision (Figure 6) indicate varied performance across Query Expansion techniques.
For our evaluation, we utilized the Ragas Context Precision metric with a top-k value of 10. Specifically, we calculated the Ragas Context Precision@10 score to assess the retrieval performance. This metric evaluates the precision of the top 10 retrieved passages in terms of their relevance to the given query. We then compared these scores across different retrieval methods to determine their effectiveness. This approach allows us to quantify and compare the retrieval accuracy of various models, providing a robust measure of their performance.

\begin{figure}[H] \centering 
\includegraphics[width=\linewidth]{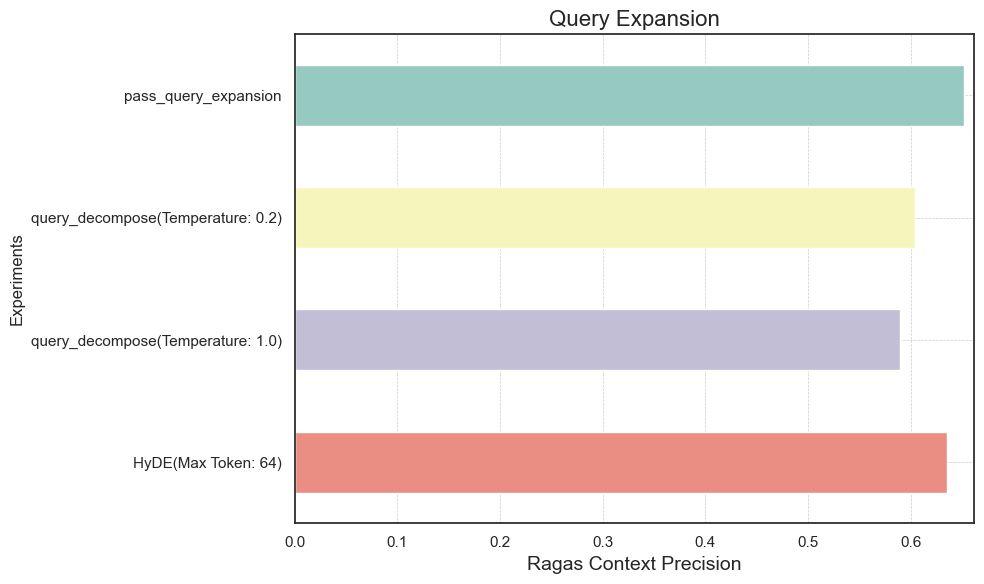} \caption{Barplots of Ragas Retrieval Precision by Query Expansion Experiments.} 
\label{fig:query-expansion} 
\end{figure}

\begin{table}[H]
\centering
\begin{tabular}{l>{\centering\arraybackslash}p{4cm}>{\centering\arraybackslash}p{5cm}}
\toprule {\textbf{Query Expansion Modules}} & \multicolumn{1}{c}{\textbf{Execution Time(seconds)}} & \multicolumn{1}{c}{\textbf{Ragas Context Precision@10}} \\
\midrule
\textbf{Pass Query Expansion} & 0.000017 & \textbf{0.651694} \\
Query Decompose(Temp: 0.2) & 0.200412 & 0.603911 \\
Query Decompose(Temp: 1.0) & 0.172025 & 0.589451 \\
HyDE(Max Token: 64) & 0.375629 & 0.634954 \\
\bottomrule
\end{tabular}
\caption{Table of Execution Time and Ragas Context Precision by Query Expansion Experiments.}
\end{table}

The highest score was achieved by pass query expansion, which uses the base query without Query Expansion. We can see that Query Expansion can improve search performance on certain data, but on other data, it can make retrieval performance worse than the base query, resulting in lower overall RAG performance. Techniques that utilize hypothetical document embedding (HyDE) show lower precision than pass and fail to improve retrieval performance. Decompose performs worse than pass and Hypothetical Document Embedding (HyDE) at both temperatures.

\subsubsection{Retrieval}

The bar plots for Context Precision (Figure 7) indicate varied performance across Retrieval techniques. 
We compared these scores across different retrieval methods to determine their effectiveness. This approach allows us to quantify and compare the retrieval accuracy of various models, providing a robust measure of their performance.

In our comparative experiment between BM25 and VectorDB, the BM25 algorithm demonstrated superior performance relative to VectorDB. 
The retrieval performance of VectorDB and BM25 differs across various datasets. Empirical experiments are necessary to determine the better approach, as semantic retrieval methods(VectorDB) sometimes outperform lexical approaches(BM25), and vice versa. In this particular dataset, BM25 demonstrated superior performance compared to traditional VectorDB methods.
Consequently, we configured the hybrid retrieval system with weights of 0.7 for BM25 and 0.3 for VectorDB to form hybrid modules. This weighting scheme was chosen to leverage the strengths of BM25 while incorporating the benefits of VectorDB, thereby optimizing the overall retrieval performance. This approach also aimed to reduce the computational cost of the experiment. However, it is important to note that alternative weight configurations can be explored to potentially enhance the performance further. Future experiments could involve varying the hybrid weights to identify the optimal balance between BM25 and VectorDB contributions.

\begin{figure}[H] \centering \includegraphics[width=\linewidth]{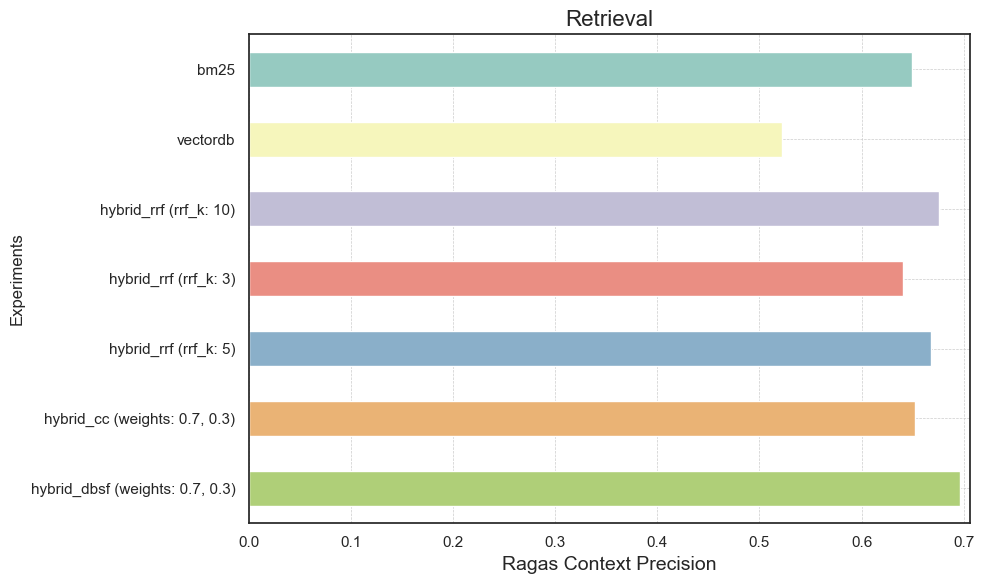} \caption{Barplots of Ragas Retrieval Precision by Retrieval Experiments.} \label{fig:query-expansion} \end{figure}

\begin{table}[H]
\centering
\begin{tabular}{l>{\centering\arraybackslash}p{4cm}>{\centering\arraybackslash}p{5cm}}
\toprule {\textbf{Retrieval Modules}} & \multicolumn{1}{c}{\textbf{Execution Time(seconds)}} & \multicolumn{1}{c}{\textbf{Ragas Context Precision@10}} \\
\midrule
Bm25 & 0.274728 & 0.649015 \\
VectorDB & 0.496673 & 0.522239 \\
Hybrid RRF (RRF-k: 10) & 0.771401 & 0.676157 \\
Hybrid RRF (RRF-k: 3) & 0.771401 & 0.640295 \\
Hybrid RRF (RRF-k: 5) & 0.771401 & 0.668342 \\
Hybrid CC (Weights: 0.7, 0.3) & 0.771401 & 0.652625 \\
\textbf{Hybrid DBSF (Weights: 0.7, 0.3)} & 0.771401 & \textbf{0.696401} \\
\bottomrule
\end{tabular}
\caption{Table of Execution Time and Ragas Context Precision by Retrieval Experiments}
\end{table}

The highest score was achieved by hybrid DBSF (weights: 0.7, 0.3), which uses the Distribution Based Score Fusion algorithm. 
The next best performance was observed with the Hybrid Reciprocal Rank Fusion (RRF) method, specifically with an RRF-k value of 10. Our analysis indicates a positive correlation between the RRF-k value and retrieval performance; as the k value increased, so did the performance metrics. 
At an RRF-k value of 3, the precision was lower than that of the baseline BM25. However, increasing the k value to 10 resulted in the second-highest performance among all tested configurations. Notably, an RRF-k value of 5 alone yielded higher precision values than Hybrid CC. 
The Hybrid CC algorithm demonstrated superior performance compared to the traditional BM25 algorithm. This suggests that, for this particular dataset, hybrid retrieval methods generally achieve higher precision than conventional retrieval methods such as BM25 and VectorDB (Vector Similarity Search).

\subsubsection{Passage Augmenter}

Passage Augmenter increases the number of Retrieved Passages, so we set top-k to 15. The mode of the prev-next augmenter is set to 'both', thereby incorporating both the previous and next paragraphs. 
In the previous retrieval step, 10 paragraphs per query are initially obtained. To enhance the context, these paragraphs are augmented by including the previous and next paragraphs, resulting in a total of 30 passages per query. The passage augmenter, configured with a top-k parameter of 15, subsequently selects the 15 highest-scoring passages from these 30 passages. These selected passages are then forwarded to the subsequent processing stage, the passage reranker.

\begin{figure}[H] \centering \includegraphics[width=\linewidth]{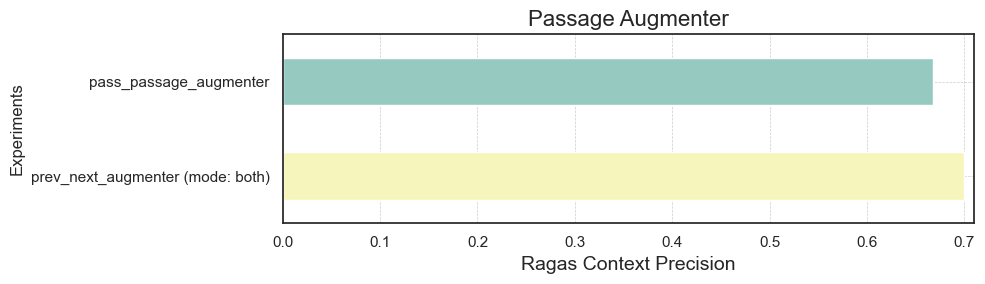} \caption{Bar plots of Ragas Retrieval Precision by Passage Augmenter Experiments.} \label{fig:query-expansion} \end{figure}

\begin{table}[H]
\centering
\begin{tabular}{l>{\centering\arraybackslash}p{4cm}>{\centering\arraybackslash}p{5cm}}
\toprule {\textbf{Passage Augmenter Modules}} & \multicolumn{1}{c}{\textbf{Execution Time(seconds)}} & \multicolumn{1}{c}{\textbf{Ragas Context Precision@15}} \\
\midrule
Pass Passage Augmenter & 0.003849 & 0.667531 \\
\textbf{Prev Next Augmenter} & 0.792790 & \textbf{0.699620} \\
\bottomrule
\end{tabular}
\caption{Table of Execution Time and Ragas Context Precision by Passage Augmenter Experiments}
\end{table}

In our experiments, the Prev Next Augmenter demonstrated superior performance compared to the Pass Passage Augmenter. This outcome is contingent on the specific corpus data used in our study. The data revealed that the preceding and succeeding passages of the retrieved passage contain valuable contextual information, which enhances the retrieval performance. This finding suggests that incorporating context from adjacent passages is more effective in leveraging contextual information than the Pass Passage Augmenter, thereby improving the overall retrieval accuracy.

\subsubsection{Passage Reranker}

The bar plots for Context Precision (Figure 9) indicate varied performance across Passage Reranker techniques.
We compared these scores across different retrieval methods to determine their effectiveness. This approach allows us to quantify and compare the retrieval accuracy of various models, providing a robust measure of their performance.

In the previous passage augmenter step, 15 paragraphs per query are obtained. However, we decided that it would be expensive to include all 15 passages in the prompt. Therefore, as a practical approach, we decided to include only 5 passages in the prompt and set the top-k parameter in the passage reranker to 5. Thus, the passage reranker calculates the scores of 15 passages per query from the previous passage augmenter. It then selects only the five passages with the highest scores and sends them to the next step, the prompt maker.

\begin{figure}[H] \centering \includegraphics[width=\linewidth]{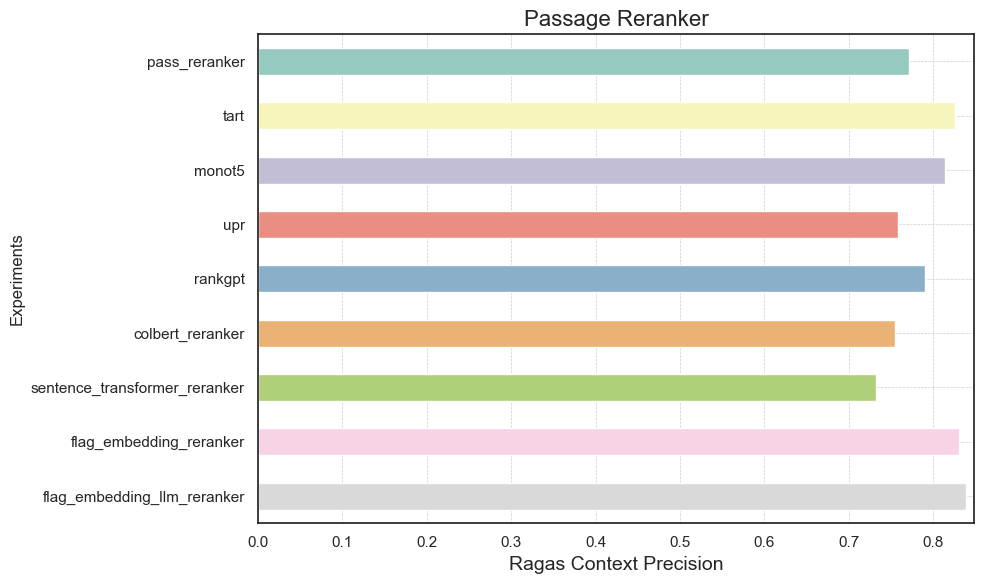} \caption{Bar plots of Ragas Retrieval Precision by Passage Reranker Experiments.} \label{fig:query-expansion} \end{figure}

\begin{table}[H]
\centering
\begin{tabular}{l>{\centering\arraybackslash}p{4cm}>{\centering\arraybackslash}p{5cm}}
\toprule {\textbf{Passage Reranker Modules}} & \multicolumn{1}{c}{\textbf{Execution Time(seconds)}} & \multicolumn{1}{c}{\textbf{Ragas Context Precision@5}} \\
\midrule
Pass Reranker & 0.000020 & 0.770846 \\
Tart & 0.282748 & 0.826207 \\
Monot5 & 1.473688 & 0.814006 \\
UPR & 0.601418 & 0.758684 \\
RankGPT & 0.219637 & 0.790732 \\
Colbert Reranker & 0.071596 & 0.755244 \\
Sentence Transformer Reranker & 0.020938 & 0.732386 \\
Flag Embedding Reranker & 0.288357 & 0.830218 \\
\textbf{Flag Embedding LLM Reranker} & 1.910619 & \textbf{0.838253} \\
\bottomrule
\end{tabular}
\caption{Table of Execution Time and Ragas Context Precision by Passage Reranker Experiments}
\end{table}

The highest performance in our evaluation was achieved by the LLM Reranker utilizing Flag Embedding. The second highest performing module was also a Flag Embedding Reranker. These results indicate that Flag Embedding rerankers are particularly effective on this dataset. Specifically, the Flag Embedding LLM Reranker, which is based on a LLM, outperformed the Flag Embedding Reranker, which is based on a language model (LM). This suggests that LLM-based rerankers are more effective than LM-based rerankers for this dataset. However, it is noteworthy that LLM-based rerankers are not universally superior. For instance, RankGPT, another LLM-based reranker, ranked fifth out of the nine experimental modules. This demonstrates variability in performance among different LLM-based rerankers.

\subsection{Generation Metric}

To evaluate the answers generated by RAG, we employed four distinct generation metrics. These metrics were applied to each experimental module across 107 queries, resulting in 107 individual accuracy scores per module. The mean value of these 107 scores was then calculated to obtain the Generation Metric Score for each module.

To facilitate a comparative analysis of the modules, we visualized the scores of the four generation metrics using bar plots. Among these metrics, METEOR, ROUGE, and Sem Score are scaled between 0 and 1, whereas G-Eval is scaled between 1 and 5. Due to the differing scales of these metrics, we plotted two separate graphs. This approach ensures a clear and accurate comparison of the performance of RAG's answers across the different experimental modules.

\subsubsection{Prompt Maker}

We utilized identical prompts for the two experimental modules, namely the f-string and the long context reorder modules. 
By using the same prompts across both modules, we ensured consistency in the experimental conditions, allowing for a more accurate comparison of their performance.
The prompts employed in these experiments were as follows:

\begin{tcolorbox}[colback=white, colframe=black, title=Prompt]
You are an expert Q\&A system that is trusted around the world for your factual accuracy.

Always answer the query using the provided context information, and not prior knowledge. Ensure your answers are fact-based and accurately reflect the context provided.

Some rules to follow:

1. Never directly reference the given context in your answer.

2. Avoid statements like 'Based on the context, ...' or 'The context information ...' or anything along those lines.

3. Focus on succinct answers that provide only the facts necessary, do not be verbose. Your answers should be max two sentences, up to 250 characters.

---------------------

(context str)

---------------------

Given the context information and not prior knowledge, answer the query.

Query: (query str)

Answer:

\end{tcolorbox}

\begin{figure}[H] \centering \includegraphics[width=\linewidth]{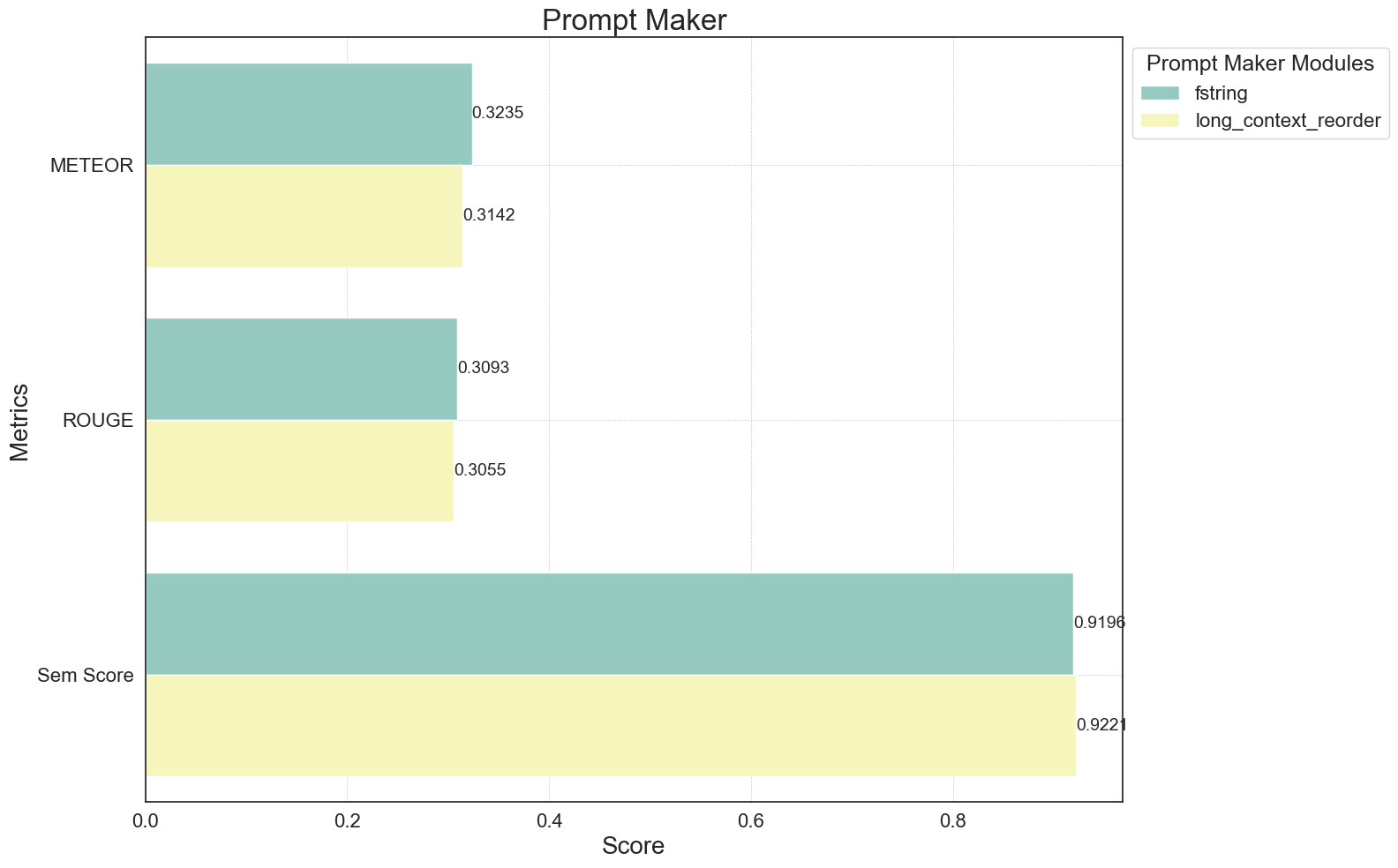} \caption{Barplots of METEOR, ROUGE and Sem Score by Prompt Maker Experiments.} \label{fig:query-expansion} \end{figure}

\begin{table}[H]
\centering
\begin{tabular}{l>{\centering\arraybackslash}p{3cm}>{\centering\arraybackslash}p{3cm}>{\centering\arraybackslash}p{3cm}}
\toprule {\textbf{Prompt Maker Modules}} & \multicolumn{1}{c}{\textbf{METEOR}} & \multicolumn{1}{c}{\textbf{ROUGE}} & \multicolumn{1}{c}{\textbf{Sem Score}} \\
\midrule
\textbf{F-string} & \textbf{0.3235} & \textbf{0.3093} & 0.9196 \\
Long Context Reorder & 0.3142 & 0.3055 & \textbf{0.9221} \\
\bottomrule
\end{tabular}
\caption{Table of METEOR, ROUGE and Sem Score by Prompt Maker Experiments}
\end{table}

For the METEOR and ROUGE metrics, the f-string module achieved slightly higher scores compared to the long context reorder module. Similarly, for the n-gram based metric, the f-string module also demonstrated superior performance.

Conversely, the long context reorder module scored marginally better on the Sem Score metric, which is based on semantic similarity. However, the differences in performance between the two modules are minimal, as illustrated in the bar plots (Figure 10).

\begin{figure}[H] \centering \includegraphics[width=\linewidth]{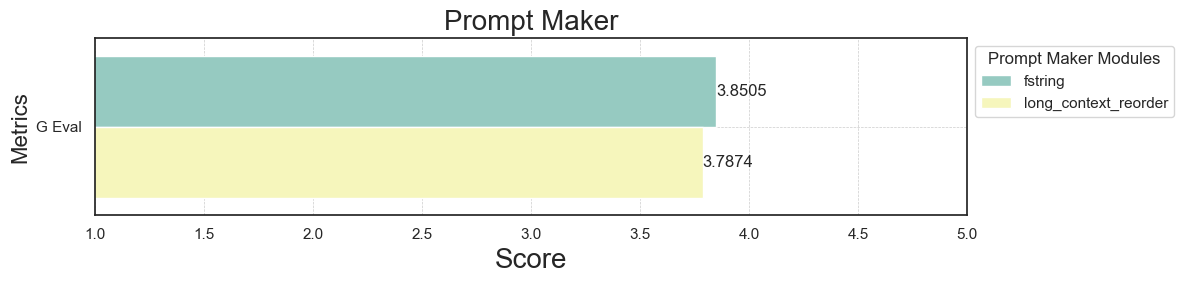} \caption{Barplots of G Eval by Prompt Maker Experiments.} \label{fig:query-expansion} \end{figure}

\begin{table}[H]
\centering
\begin{tabular}{l>{\centering\arraybackslash}p{3cm}}
\toprule {\textbf{Prompt Maker Modules}} & \multicolumn{1}{c}{\textbf{G Eval}}\\
\midrule
\textbf{F-string} & \textbf{3.8505} \\
Long Context Reorder & 3.7874 \\
\bottomrule
\end{tabular}
\caption{Table of G Eval by Prompt Maker Experiments}
\end{table}

In the evaluation using the G-Eval metric, which is judged by an LLM, the f-string module outperformed the long context reorder module. 

\subsubsection{Generator}

The LLM model used in our experiments was fixed to GPT-3.5-turbo with a temperature setting of 0.0. As the experiment was not designed to compare the performance between different LLM models, we do not include graphs comparing performance results across various LLM models.

Below are the final results obtained using the selected model and pipeline:

\begin{table}[H]
\centering
\begin{tabular}{l>{\centering\arraybackslash}p{2.5cm}>{\centering\arraybackslash}p{2.5cm}>{\centering\arraybackslash}p{2.5cm}>{\centering\arraybackslash}p{2.5cm}}
\toprule {\textbf{LLM}} & \multicolumn{1}{c}{\textbf{METEOR}} & \multicolumn{1}{c}{\textbf{ROUGE}} & \multicolumn{1}{c}{\textbf{Sem Score}} & \multicolumn{1}{c}{\textbf{G Eval}} \\
\midrule
\textbf{gpt-3.5-Turbo} & \textbf{0.3246} & \textbf{0.3054} & \textbf{0.9186} & \textbf{3.8037} \\
\bottomrule
\end{tabular}
\caption{Table of Generation metrics by Generator Experiments}
\end{table}

In this experiment, both the Prompt Maker and the generator module utilized the same model (GPT-3.5-turbo) and identical temperature settings (0.0). the performance values of the generator module were directly influenced by the outcomes of the Prompt Maker. As a result,  Although the settings for both modules were identical, the inherent variability of the LLM resulted in different metric values. While the inherent variability in LLM outputs led to slight variations in the Generation Metric values, these differences were minimal. This indicates that both modules performed consistently under the same conditions, making the performance values of the two modules directly comparable.

\section{Discussion}

The lower performance with the query expansion method can be attributed to the fact that the queries in the configured evaluation dataset were not multi-hop.
Hybrid DBSF showed the highest performance among retrieval methods, and the flag embedding llm reranker showed the highest performance among rerankers.
The use of the prev-next augmenter as a passage augmenter slightly improved retrieval performance. Also, not using the long context reorder resulted in slightly better performance than using it.

At reranker node, some rerankers, such as UPR, ColBERT, and Sentence Transformer, performed worse than the baseline Pass Reranker without reranking. 
This indicates that, depending on the dataset, reranking can sometimes degrade performance rather than improve it. The lowest performance was observed with the Sentence Transformer Reranker.
The ARAGOG dataset combines multiple papers to form a single query, which may require knowledge from several passages to generate an answer. However, UPR generates queries based on a single passage, likely leading to its lower performance.
ColBERT is embedding based re-ranking method. In retrieval stage result, the  \(ragas\_context\_precision@10\) of BM25(sparse retrieval) is 0.649015 and, the \(ragas\_context\_precision@10\) of VectorDB(dense retrieval). The ColBERT method is conceptually similar to Dense retrieval. It may be suggested that the language model has limited semantic understanding of the domain-specific dataset. especially, Sentence Transformer Reranker use \(ms-marco-MiniLM-L-2-v2\) model. Since it has smaller model parameter size than the models used in other rerankers, semantic understanding issues arising from the domain-specific data may be more pronounced.

\section{Limitations}
\begin{itemize}
\item \textbf{Normalization Methods}: There are various normalization methods available when executing hybrid cc retrieval, such as TMM. (\cite{hybrid-cc}) In this paper, we computed hybrid retrieval using only two different normalization methods. The performance of other normalization methods can vary.

\item \textbf{Small search space of hybrid retrieval parameter}: In the hybrid retrieval method, both cc and rrf, the hyper-parameters can affect the performance of retrieval. However, in this paper, we evaluated only a few hyper-parameters due to the high cost of retrieval metrics.

\item \textbf{Expensive to reproduce}: The high computational requirements of RAGAS retrieval metrics make it difficult to reproduce experiments, posing a barrier to the validation and verification of results.

\item \textbf{Lack of Meta-Evaluation}: There is no meta-evaluation process to quantify how much AutoRAG improves the RAG pipeline performance compared to other methods.

\item \textbf{Inherent LLM variability}: LLMs are known to produce slightly different outputs even when provided with the same input, due to their stochastic nature. Therefore, the results and interpretations of the pipeline may not be entirely robust.

\end{itemize}

\section{Conclusion}

In this paper, we use AutoRAG, an automated RAG optimization framework. It identifies the best module for each node using a greedy approach, and the combination of these modules is expected to achieve performance close to the optimal RAG combination.

Using AutoRAG, we were able to automatically optimize the RAG system for the dataset from ARAGOG(\cite{eibich2024aragogadvancedragoutput}).

To facilitate future RAG optimization and RAG system evaluation using various datasets and additional RAG techniques, we have released the AutoRAG framework as code on the Github repository. By utilizing AutoRAG, it will be possible to conduct relative evaluations of RAG techniques not covered in this paper, as well as validations using datasets with a larger number of QA pairs from various domains.

\section{Future Work}

\begin{itemize}
\item \textbf{Evaluation of AutoRAG Optimization Capabilities}: Evaluating the AutoRAG methodology itself is not an easy task. However, it is important to assess the performance of the AutoML framework itself. A performance comparison with methodologies like AutoRAG-HP (\cite{autorag-hp}) would also be necessary.

\item \textbf{Experiments on More Diverse Datasets}: Since automatic testing is possible using AutoRAG, experiments on more datasets should be conducted. This will help understand the characteristics of datasets from various domains and gather information on suitable RAG techniques.

\item \textbf{Experiments on Additional RAG Modules}: In RAG systems, it is crucial to set appropriate chunking strategies and use suitable parsing techniques for documents. Additionally, techniques like Modular RAG(\cite{gao2024modularragtransformingrag}) are being applied. Modular RAG is controlled by multiple components for entire RAG process, then the RAG pipeline can be more flexible and complex than simple linear RAG pipeline. Optimization and experiments can be conducted, including these RAG techniques.
\end{itemize}

\bibliography{references}

\end{document}